\newif\ifanonymous   \anonymousfalse
\newif\ifabstract    \abstracttrue

\documentclass[11pt]{article}

\usepackage{booktabs}
\usepackage{graphicx}
\usepackage{amsmath}
\usepackage{amssymb}
\usepackage[margin=1in]{geometry}
\usepackage{caption}
\usepackage{hyperref}
\usepackage{enumitem}
\usepackage[style=apa,backend=biber]{biblatex}
\usepackage[T1]{fontenc}
\usepackage[utf8]{inputenc}
\usepackage{tabularx}
\usepackage{threeparttable}
\usepackage{adjustbox}
\usepackage{xcolor}
\usepackage{setspace}
\usepackage{float}
\usepackage{rotating}
\usepackage{placeins}

\hypersetup{
  colorlinks  = true,
  linkcolor   = blue!70!black,
  citecolor   = blue!70!black,
  urlcolor    = blue!70!black,
  pdfborder   = {0 0 0}
}

\ifanonymous
  \title{\textbf{\Large Researchers waste 80\% of LLM annotation costs by classifying one text at a time}}
\else
  \title{\textbf{\Large Researchers waste 80\% of LLM annotation costs by classifying one text at a time}\textsuperscript{\dag}}
\fi

\ifanonymous
  \author{}
\else
  \author{
    Christian Pipal\textsuperscript{1,*},
    Eva-Maria Vogel\textsuperscript{1},
    Morgan Wack\textsuperscript{1},
    Frank Esser\textsuperscript{1} \\
    \vspace{6pt}
    \small\textsuperscript{1}Department of Communication and Media Research (IKMZ), University of Zurich \\
    \small\textsuperscript{*}Corresponding author, \href{mailto:c.pipal@ikmz.uzh.ch}{c.pipal@ikmz.uzh.ch}
  }
\fi

\date{}

\begin{document}

\maketitle

\ifanonymous\else
\begingroup
\renewcommand{\thefootnote}{\dag}
\footnotetext{C.P.\ designed research, performed research, analyzed data, and wrote the paper. E.-M.V., M.W., and F.E.\ contributed to research design and edited the paper. The authors declare no competing interest.}
\endgroup
\setcounter{footnote}{0}
\fi

\ifabstract
\begin{abstract}
\noindent
Large language models (LLMs) are increasingly being used for text classification across the social sciences, yet researchers overwhelmingly classify one text per variable per prompt. Coding 100,000 texts on four variables requires 400,000 API calls. Batching 25 items and stacking all variables into a single prompt reduces this to 4,000 calls, cutting token costs by over 80\%. Whether this degrades coding quality is unknown. We tested eight production LLMs from four providers on 3,962 expert-coded tweets across four tasks, varying batch size from 1 to 1,000 items and stacking up to 25 coding dimensions per prompt. Six of eight models maintained accuracy within 2 pp of the single-item baseline through batch sizes of 100. Variable stacking with up to 10 dimensions produced results comparable to single-variable coding, with degradation driven by task complexity rather than prompt length. Within this safe operating range, the measurement error from batching and stacking is smaller than typical inter-coder disagreement in the ground-truth data.
\end{abstract}

\vspace{0.5em}
\fi

\setcounter{page}{1}
\doublespacing
\setlength{\parskip}{0pt}
\setlength{\parindent}{1.5em}

\section*{Introduction}

Large language models (LLMs) have quickly become standard tools for automated content analysis across the social sciences \parencite{bail2024generativeai}, and are now used to classify millions of social media posts \parencite{hartman2025negative}, code news content at scale \parencite{pew2024influencers}, and score hundreds of thousands of academic texts \parencite{theorysociety2026ideology}. Validation work has established that LLMs match or exceed human coders on tasks central to this research \parencite{gilardi2023chatgpt, tornberg2025outperform, ornstein2025stochastic}. A methodological literature has emerged to guide prompt design and model selection \parencite{baumann2025hacking, carlson2026annotate}. Yet two implementation choices that shape every LLM-based project remain untested.

Item batching determines how many texts are grouped in a single prompt, while variable stacking determines how many coding dimensions are requested per call. Published practice is nearly universal in its conservatism, with one text, one variable, and one prompt per API call \parencite{gilardi2023chatgpt, tornberg2025outperform, heseltine2024substitute}. Coding 100,000 texts on four variables at this baseline requires 400,000 API calls. Batching 25 items and stacking all four variables reduces this to 4,000 calls with over 80\% cost savings, because instructions are transmitted once rather than repeated per call. For projects of the scale described above, these choices can determine the feasibility of a study.

The computer science literature shows that batching is viable at small sizes on standard NLP benchmarks \parencite{cheng2023batch, lin2024batchprompt}, and that multi-task inference is architecture-dependent \parencite{son2024multitask, dagostino2025degradation}. These findings do not address codebook-driven classification typical in social science content analysis. Whether batching and stacking degrade quality in this setting, and whether their effects compound, is an open question. We provide the first systematic test, varying batch size from 1 to 1,000 and stacking up to 25 variables across eight production LLMs and 960,000 classifications.

\section*{Results}

\begin{figure}[H]
\centering
\begin{minipage}[t]{0.48\textwidth}
\centering
\textbf{A: Item Batching}\\[0.3em]
\includegraphics[width=\textwidth]{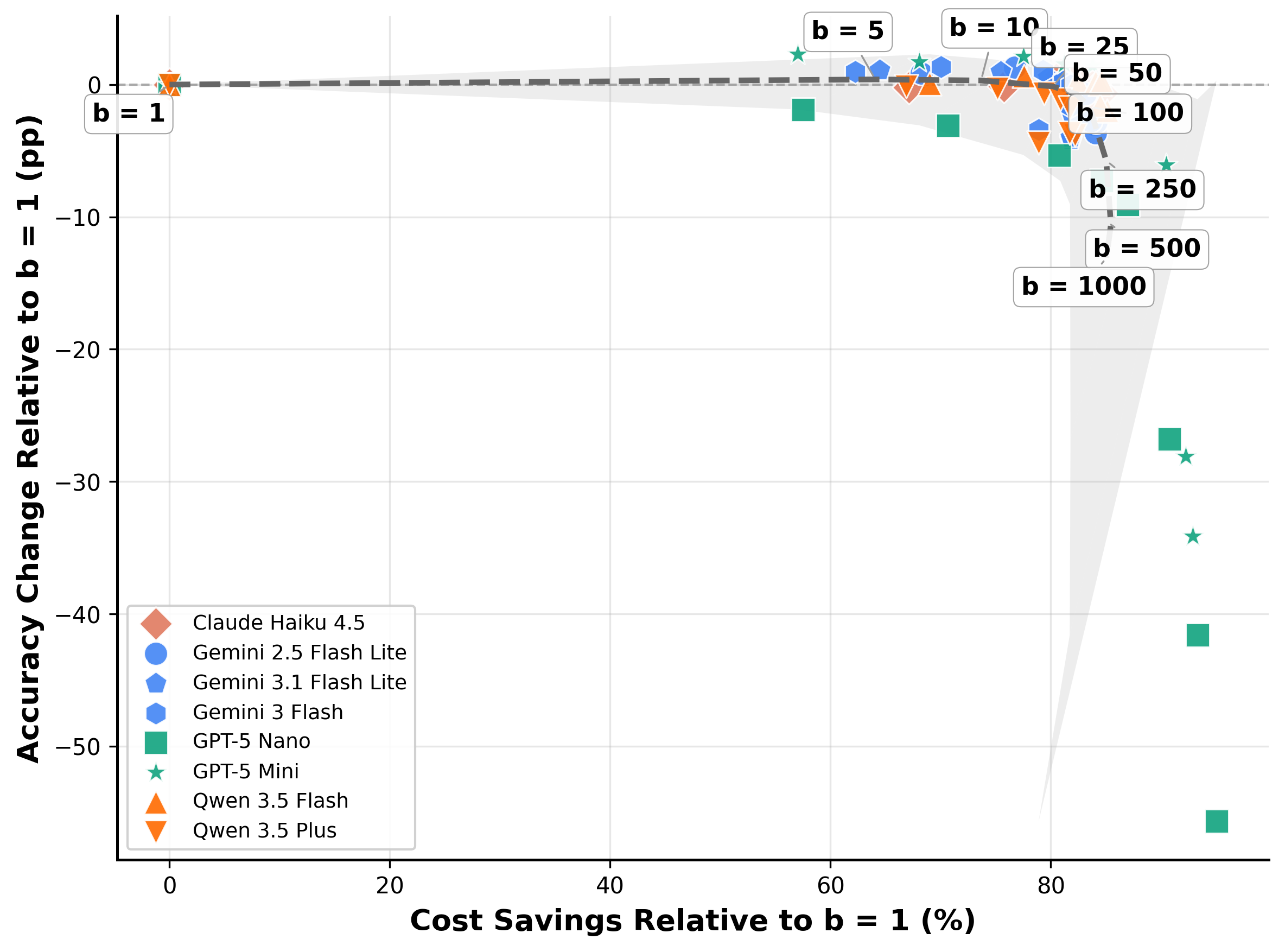}
\end{minipage}
\hfill
\begin{minipage}[t]{0.48\textwidth}
\centering
\textbf{B: Variable Stacking}\\[0.3em]
\includegraphics[width=\textwidth]{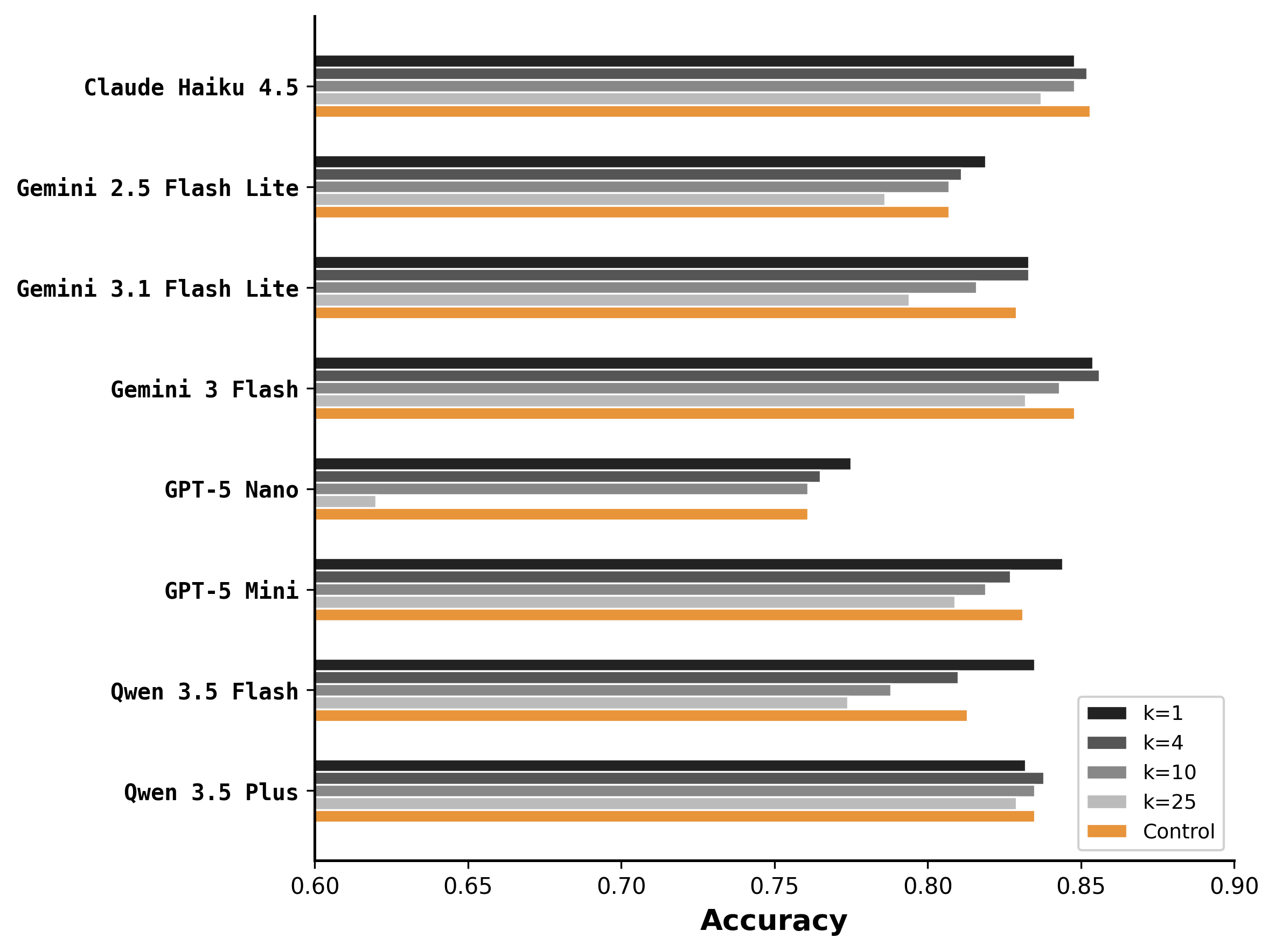}
\end{minipage}
\caption{Batching and stacking are safe within bounds. (\textit{A}) Cost savings (\%) vs.\ accuracy change (pp) relative to $b = 1$ for each model at each batch size. The dashed line traces the mean trajectory across all eight models; the shaded region marks the min--max range. Batch size labels mark progression along this trajectory. Six models remain near zero accuracy loss through $b = 100$ (${\sim}84\%$ cost savings). Two OpenAI reasoning models (green) collapse at $b \geq 250$, driving the shaded region downward. (\textit{B}) Overall accuracy (mean across four variables) under stacking conditions at $b = 25$. $k = 1$: single-variable baseline from Study~1. $k = 4, 10, 25$: number of simultaneous coding dimensions in Study~2. Control (orange): same prompt length as $k = 25$ but only four variables plus filler text, showing that degradation reflects task complexity, not prompt length.}
\label{fig:main}
\end{figure}

\textbf{Item batching.} We LLM-coded 3,962 expert-labeled tweets \parencite{gilardi2023chatgpt} on four variables (relevance, problem/solution, stance, topic) across nine batch sizes ($b \in \{1, 5, 10, 25, 50, 100, 250, 500, 1000\}$) using eight LLMs from four providers. Accuracy remained stable for most models through $b = 100$ (Fig.~\ref{fig:main}A). Six of eight models showed accuracy drops below 2 pp between $b = 1$ and $b = 100$ while saving over 80\% in token costs. Claude Haiku 4.5 was the most robust, with a maximum drop of 1.1 pp across all batch sizes. Degradation at $b \geq 250$ was model-specific. Google and Alibaba models declined 3--4 points by $b = 1000$. Two OpenAI reasoning models (GPT-5-nano, GPT-5-mini), which cannot run at temperature 0, collapsed at these sizes, losing 27--36 points with substantial parse failures. No other model exceeded 1.5\% parse failure at any batch size.

\begin{figure}[H]
\centering
\includegraphics[width=\textwidth]{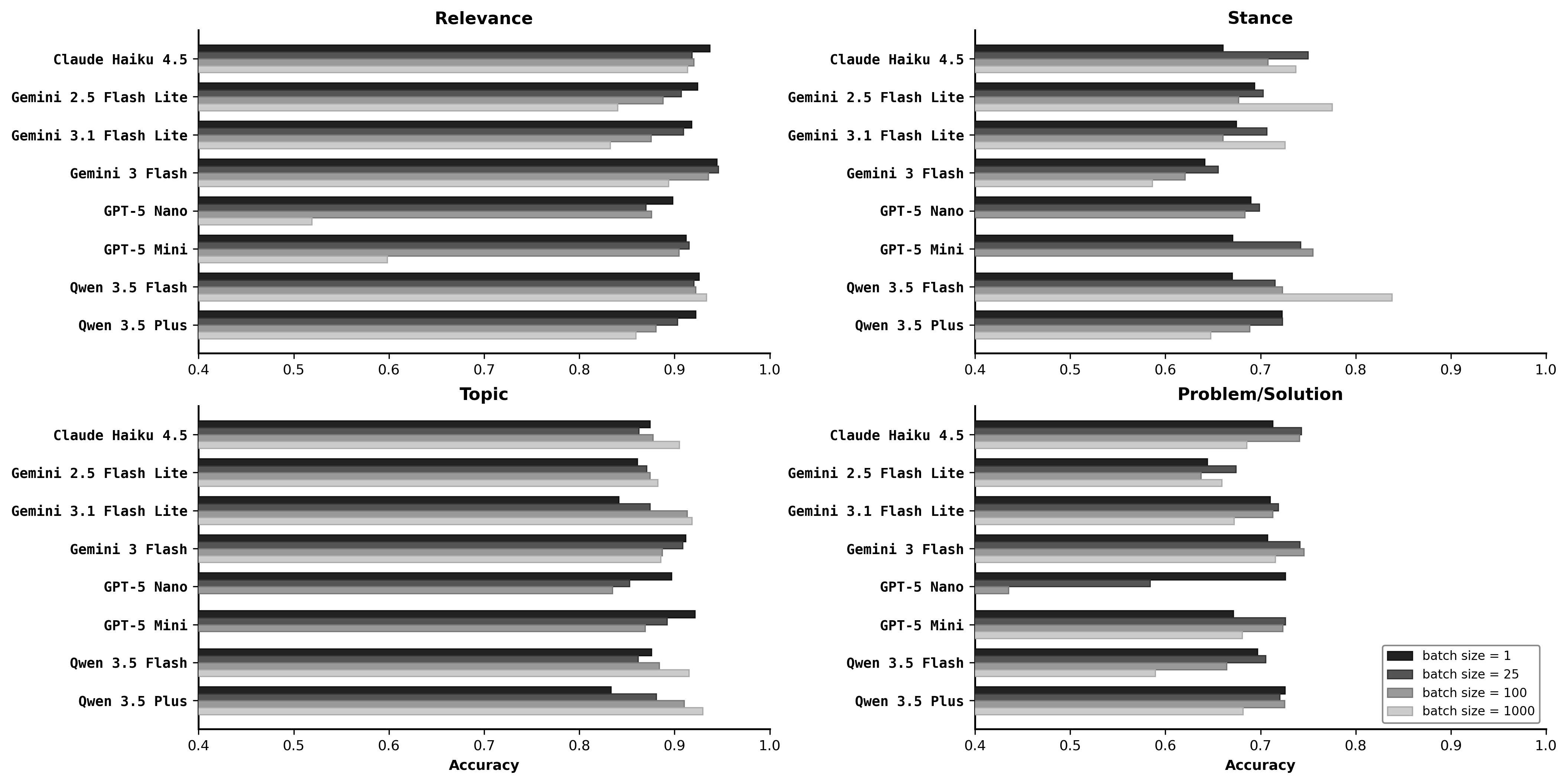}
\caption{Accuracy by model, variable, and batch size (Study~1). Each panel shows one of the four variables. Bars represent batch sizes from $b = 1$ (darkest) to $b = 1000$ (lightest). Six of eight models maintain stable accuracy through $b = 100$ across all four variables. GPT-5 Nano shows progressive degradation starting at $b = 5$. Topic accuracy increases with batch size for several models, possibly reflecting an auto-demonstration effect in which co-presented tweets provide implicit distributional information.}
\label{fig:detail}
\end{figure}

The four variables responded differently to batching (Fig.~\ref{fig:detail}). Relevance (binary, $N = 3{,}179$) was the most sensitive in relative terms, with most models losing 5--9 pp by $b = 1000$, though absolute accuracy remained above 83\% for all non-OpenAI models. Problem/solution (three-class) showed moderate degradation of 2--5 points. Stance (three-class, $N = 783$) was noisy throughout due to class imbalance. Topic (six-class, $N = 611$) showed a different pattern. Accuracy \textit{increased} at larger batch sizes for several models, possibly because co-presented tweets provide distributional information that helps calibrate category boundaries.

\textbf{Variable stacking.} We crossed three levels of stacking ($k \in \{4, 10, 25\}$ simultaneous coding dimensions) with batch size $b = 25$, comparing against the single-variable baseline ($k = 1$) from Study~1 (Fig.~\ref{fig:main}B). Moving from $k = 1$ to $k = 25$ reduced accuracy by 1.6--6.1 pp for seven of eight models, though GPT-5-nano dropped 15.5 points. Stacking up to $k = 10$ produced results within 2.2 pp of the single-variable baseline for all models except GPT-5-nano. The batching-stacking interaction was additive rather than multiplicative. Accuracy at $b = 250$ showed comparable stacking effects, so researchers can optimize along both dimensions without a compounding penalty.

\textbf{Prompt length vs.\ task complexity.} A control prompt matching the token length of the $k = 25$ condition but containing only four ground-truth variables (with background text filling the remaining space) performed at the level of $k = 4$ for all models (Fig.~\ref{fig:main}B, orange bars). Degradation under stacking is caused by the demand of managing multiple classification schemes simultaneously, not by additional tokens in the prompt.

\section*{Discussion}

Batch sizes of 25--100 are safe for most current LLMs, saving over 80\% in costs with accuracy loss below 2 pp. Stacking up to 10 coding dimensions is also viable, with degradation driven by task complexity rather than prompt length. Model choice matters more than batch size. At any given $b$, the gap between the best and worst model exceeds the gap across batch sizes for any given model.

A researcher coding 100,000 tweets on four variables with a mid-tier model can reduce costs by an estimated ${\sim}$\$337 to under \$50 (or under \$10 with a budget model) by batching at $b = 25$ and stacking all variables. The measurement error introduced within this safe range is smaller than the inter-coder disagreement in the ground-truth data itself between 6--21\%), which indicates that these efficiency gains fall within the tolerance the field already accepts for human-coded data \parencite{teblunthuis2024misclassification}.

Batch size and stacking configuration are researcher degrees of freedom that should be reported alongside model name and prompt design \parencite{baumann2025hacking, carlson2026annotate}. Our results are based on short English-language tweets from a single domain. Whether these results generalize to longer documents, other languages, or more complex tasks requires further investigation.

\section*{Materials and Methods}

\textbf{Data.} We used replication data from \textcite{gilardi2023chatgpt}, compromising 3,962 tweets about content moderation posted between Jan. 2020 and Apr. 2021. Each tweet was coded by two trained research assistants, and disagreements were adjudicated to produce ground-truth labels on four variables: relevance ($N = 3{,}179$, binary), problem/solution ($N = 1{,}500$, three-class), stance on Section 230 ($N = 783$, three-class), and topic ($N = 611$, six-class). Human inter-coder reliability (Cohen's $\kappa$) ranged from 0.63 (stance) to 0.88 (relevance). Not all tweets are coded on all variables because the original cascading annotation design means sample sizes differ across variables.

\textbf{Models.} We tested eight production LLMs from four providers: Claude Haiku 4.5 (Anthropic), GPT-5-mini and GPT-5-nano (OpenAI), Gemini 3 Flash, 3.1 Flash-Lite, and 2.5 Flash-Lite (Google), and Qwen 3.5 Plus and 3.5 Flash (Alibaba). All models ran at temperature 0 except GPT-5-mini and GPT-5-nano, which are locked at temperature 1 (no 0 option).

\textbf{Prompt design.} Study 1 prompts code a single variable per prompt and include two few-shot examples per class drawn from the \textcite{gilardi2023chatgpt} codebook. Study 2 prompts code multiple variables simultaneously using 28 shared worked examples (the union of all Study 1 examples), labeled on all dimensions. Dummy variables at k=10 and k=25 are plausible but unevaluated tasks (e.g., sentiment, sarcasm). All prompts instruct the model to return a JSON array of labels.

\textbf{Experimental design.} Study~1: $9$ batch sizes $\times$ $4$ variables $\times$ $8$ models $= 288$ conditions. Study~2: $3$ stacking levels $\times$ $2$ batch sizes ($b = 25$, $b = 250$) $\times$ $8$ models $= 48$ conditions, plus position variants (ground-truth variables at beginning, middle, or end of codebook) and a prompt-length control. Total: 376 conditions, 963,151 individual classifications. The dataset was shuffled once (seed=42) and held constant across conditions.

\textbf{Analysis.} Accuracy (proportion matching ground truth) with 95\% bootstrap CIs (500 resamples).

\textbf{AI use disclosure.} LLMs are the object of study in this paper. Claude Sonnet 4.6 (Anthropic) was used for language editing. All analytical decisions are the authors'.

\subsection*{Data availability statement}
Replication data from \textcite{gilardi2023chatgpt} are available via the Harvard Dataverse. All prompts, raw model outputs, parsed results, and analysis code are available at \url{https://osf.io/fmyvz/}.


\printbibliography

\clearpage
\appendix
\pagenumbering{arabic}
\setcounter{page}{1}
\renewcommand{\thetable}{\Alph{section}.\arabic{table}}
\renewcommand{\thefigure}{\Alph{section}.\arabic{figure}}

\begin{singlespace}

\section{Supporting Information}

\subsection*{Extended Methods}

\subsubsection*{Data}

The evaluation data come from the replication materials of \textcite{gilardi2023chatgpt}, made available via the Harvard Dataverse. The corpus consists of tweets related to content moderation on social media, originally sampled from the Twitter API between January 2020 and April 2021. The data were re-used by \textcite{alizadeh2025opensource} to compare multiple LLMs on the same classification tasks. Our evaluation dataset contains 3,962 tweets with human-coded ground-truth labels across four classification variables.

Two trained research assistants independently coded each variable. Disagreements were adjudicated to produce the final ground-truth labels. Inter-coder reliability (Cohen's $\kappa$) ranged from 0.63 for stance to 0.88 for relevance.

Not all tweets are coded on all variables. The original annotation used a cascading design: only tweets deemed relevant to content moderation were further coded for problem/solution and topic. Stance labels come from a separate annotation campaign focused on Section 230 tweets, with zero overlap with the other variables.

Relevance is moderately balanced (59\% relevant, 41\% irrelevant). Problem/solution is well balanced across three classes (30/32/38\%). Stance has a severely imbalanced ``in favor'' class with only 35 items (4.5\%), which produces unreliable per-class F1 estimates. Topic has six classes, but ``Twitter Support'' contains only 6 items and ``Other'' has zero items in the evaluation set.

\subsubsection*{Models}

We tested eight production LLMs from four providers:

Anthropic: Claude Haiku 4.5 (\$1.00/\$5.00 per million input/output tokens).

OpenAI: GPT-5-mini (\$0.25/\$2.00), GPT-5-nano (\$0.20/\$1.25). Both are reasoning models with hidden chain-of-thought; temperature is locked at 1.

Google: Gemini 3 Flash (\$0.50/\$3.00), Gemini 3.1 Flash-Lite (\$0.25/\$1.50), Gemini 2.5 Flash-Lite (\$0.10/\$0.40).

Alibaba: Qwen 3.5 Plus (\$0.40/\$2.40), Qwen 3.5 Flash (\$0.10/\$0.40).

All models (except GPT-5-mini and GPT-5-nano) were run at temperature 0 to maximize determinism with reasoning turned off. GPT-5-mini and GPT-5-nano were run at the default temperature 1 (which cannot be set to 0), and the reasoning parameter set to ``low'' (which cannot be turned off). A compliance system prompt was added to GPT-5-nano after initial runs revealed that it would sometimes refuse classification at large batch sizes, interpreting the request as outside its intended use case. The exact system prompt text is included in the replication package.

\subsubsection*{Prompt Design}

Study 1 prompts code a single variable per prompt. Each includes two few-shot examples per class drawn from the \textcite{gilardi2023chatgpt} codebook, yielding 4 examples for relevance, 6 for problem/solution, 6 for stance, and 12 for topic. The model is instructed to return a JSON array of labels in the order presented. The same examples are used across all batch sizes and models.

Study 2 prompts code multiple variables simultaneously. All Study 2 prompts share the same 28 worked examples (the union of all Study 1 few-shot tweets), each labeled on all dimensions present in that prompt. Study 2 uses three levels of variable stacking: $k = 4$ (the four ground-truth variables only), $k = 10$ (four ground-truth plus six plausible dummy variables such as sentiment, sarcasm, and call to action), and $k = 25$ (four ground-truth plus 21 dummy variables). The dummy variables are not evaluated; they serve only to increase the number of simultaneous coding dimensions. Approximate prompt token counts (excluding tweet text): $k = 4 \approx 2{,}300$ tokens, $k = 10 \approx 3{,}300$ tokens, $k = 25 \approx 5{,}700$ tokens, Control $\approx 5{,}700$ tokens.

Position variants: For $k = 10$ and $k = 25$, three orderings were tested. In the ``beginning'' condition, the four ground-truth variables appear first in the codebook. In ``middle,'' they are placed in the center. In ``end,'' they appear last. All three variants use the same 28 worked examples with JSON fields reordered accordingly. The results of position variants are presented in the replication package.

Control condition: The control prompt contains only the four ground-truth variables but includes a block of background text on the history and regulation of content moderation, padding the prompt to the same token length as $k = 25$. This design isolates the effect of prompt length from the effect of task complexity.

The complete prompt text for all conditions is available in the replication package at \url{https://osf.io/fmyvz/}.

\subsubsection*{Experimental Design}

Study 1 varies the number of items per prompt across nine levels: $b \in \{1, 5, 10, 25, 50, 100, 250, 500, 1000\}$. Each of the four classification variables is coded separately with its own prompt. This yields $9 \times 4 \times 8 = 288$ conditions.

Study 2 crosses three levels of variable stacking ($k \in \{4, 10, 25\}$) with two batch sizes ($b \in \{25, 250\}$), yielding $3 \times 2 \times 8 = 48$ conditions. Position tests at $b = 25$ add $2 \times 2 \times 8 = 32$ conditions. The control condition adds 8 conditions (one per model at $b = 25$).

The total experiment comprises 376 conditions and produced 963,151 individual classifications across 96,065 API calls at a total cost of approximately \$152.

For each condition, the full dataset was shuffled with a fixed random seed (seed $= 42$) and partitioned into batches of size $b$. The same shuffled order was held constant across all conditions so that the only source of variation is batch size and stacking level.

\subsubsection*{Analysis}

The primary metric is accuracy (proportion of items matching the ground-truth label). Accuracy is the natural metric for comparing conditions applied to the same dataset where class distributions are constant. Macro F1, which penalizes poor performance on minority classes, produces consistent patterns but is noisier due to the severely imbalanced stance variable.

All confidence intervals are 95\% bootstrap intervals based on 500 resamples. For Study 1, degradation is assessed by comparing accuracy at each batch size to the $b = 1$ baseline. For positional effects within batches, items are divided into quintiles by position and accuracy is compared across quintiles. For Study 2, accuracy in multi-variable conditions is compared to the single-variable baseline from Study 1 at the same batch size.

Each experimental condition is run once with a single shuffle order. At extreme batch sizes ($b = 500$, $b = 1000$) for small-$N$ variables such as stance ($N = 783$) and topic ($N = 611$), accuracy estimates rest on one or two API calls.

\subsection*{Additional Results}

Full accuracy tables by model and batch size, macro F1 scores, bootstrap confidence intervals, per-variable accuracy under stacking conditions, positional accuracy within batches, parse failure rates, token usage analysis, and individual model accuracy plots are available in the replication package at \url{https://osf.io/fmyvz/}.

\FloatBarrier

\end{singlespace}

\end{document}